\pdfoutput=1

\documentclass[11pt]{article}

\usepackage{acl}

\usepackage{times}
\usepackage{latexsym}

\usepackage[T1]{fontenc}

\usepackage[utf8]{inputenc}
\usepackage{CJK}

\usepackage[whole]{bxcjkjatype}

\usepackage{microtype}
\usepackage{graphicx}
\usepackage{caption}
\usepackage{subcaption}
\usepackage{comment}
\usepackage{pmboxdraw}
\DeclareUnicodeCharacter{2581}{\pmboxdrawuni{2581}}

\title{Tagged End-to-End Simultaneous Speech Translation Training \\ using Simultaneous Interpretation Data}

\author{Yuka Ko \quad Ryo Fukuda \quad Yuta Nishikawa \quad Yasumasa Kano\\
\textbf{Katsuhito Sudoh \quad Satoshi Nakamura}\\
  Nara Institute of Science and Technology \\
  \texttt{ko.yuka.kp2@is.naist.jp}
  }

\begin{document}
\maketitle
\begin{abstract}
Simultaneous speech translation (SimulST) translates partial speech inputs incrementally.
Although the monotonic correspondence between input and output is preferable for smaller latency, it is not the case for distant language pairs such as English and Japanese.
A prospective approach to this problem is to mimic simultaneous interpretation (SI) using SI data to train a SimulST model.
However, the size of such SI data is limited, so the SI data should be used together with ordinary bilingual data whose translations are given in offline.
In this paper, we propose an effective way to train a SimulST model using mixed data of SI and offline.
The proposed method trains a single model using the mixed data with style tags that tell the model to generate SI- or offline-style outputs.
Experiment results show improvements of BLEURT in different latency ranges, and our analyses revealed the proposed model generates SI-style outputs more than the baseline.
\end{abstract}

\section{Introduction}
Simultaneous speech translation (SimulST) is a technique to translate speech incrementally without waiting for the end of a sentence.
Since SimulST should work in small latency against the input speech, monotonic translation following the word order of the source language is preferable.
However, making translation monotonic is not trivial
especially for distant language pairs with different word orders, such as English and Japanese.
Most recent SimulST studies still use parallel corpora only with offline translations and potentially have the limitation to work in a monotonic way.

A prospective approach to this problem is to use SI data to train a SimulST model for mimicking simultaneous interpretation (SI).
There are several SI data resources developed so far for English-Japanese~\cite{Tohyama-et-al-2004, shimizu2013constructing, doi-etal-2021-large}. 
Despite these efforts, SI data are still very small compared to bilingual data based on offline translations.
Using such scarce SI data to fine-tune an offline translation model causes overfitting on the small SI data. 
Training a model using mixed data of offline and SI data is another option to mitigate the problem of data scarcity, but the simple data mixture causes confusion between the output styles of offline translation and SI.

In this paper, we propose a method to train a SimulST model using mixed data of SI and offline translation with style tags to tell the model to generate SI- or offline-style output selectively. 
It has the advantage of sharing two different styles in a single model and generating SI-style outputs by putting the SI-style tag in the decoding, which are leveraged by offline translation data.
Experiment results using MuST-C and small SI data showed improvements of BLEURT by the proposed method over the baselines in different latency ranges. Further analyses revealed that the proposed model generates more appropriate SI-style outputs than baselines.

\begin{figure*}[t]
    \centering
    \includegraphics[width=0.9\linewidth]{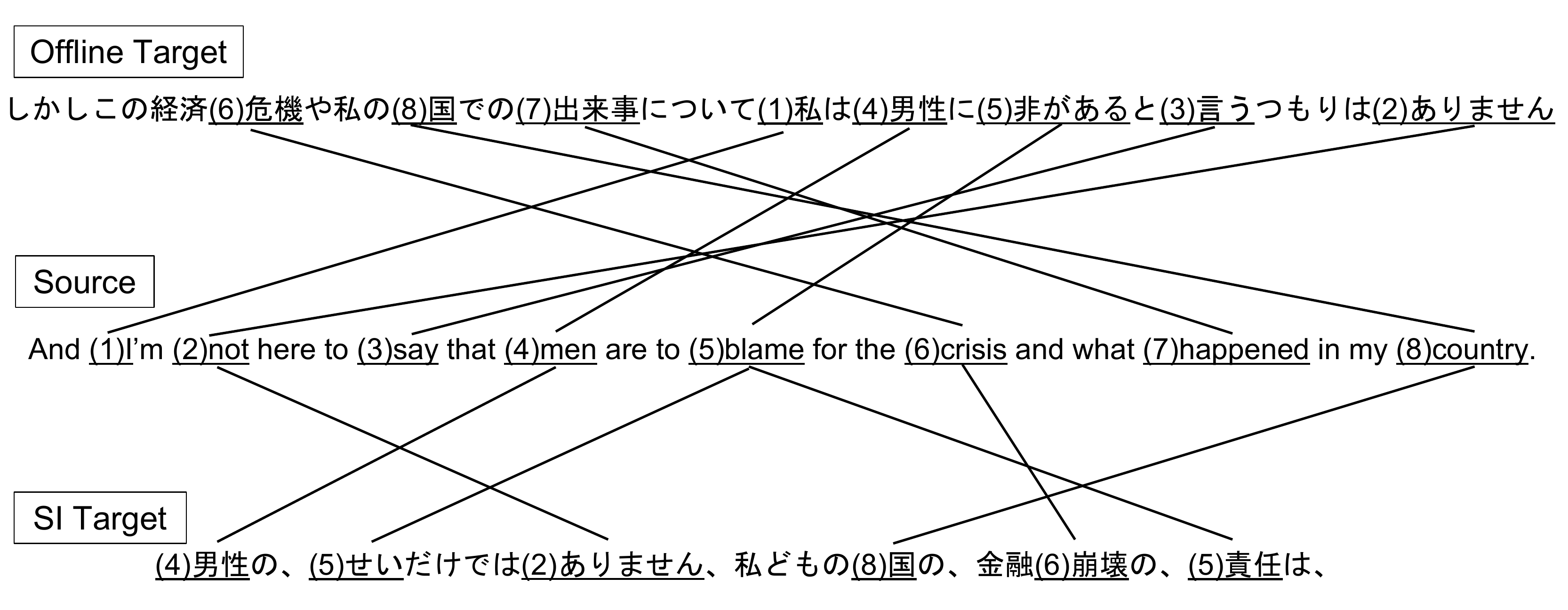}
    \caption{Example of English-to-Japanese offline translation and SI.}
    \label{tab:example-sentence-in-si-off}
\end{figure*}

\section{Related Work}
There have been many studies on simultaneous translation for text and speech in decades~\cite{fugen2007simultaneous, oda2014optimizing, dalvi-etal-2018-incremental}.
Most recent approaches are based on deep neural networks and have evolved with the technologies of neural machine translation (NMT)~\cite{gu-etal-2017-learning} and neural speech recognition (ASR)~\cite{rao2017exploring}.
An important advantage of the neural SimulST methods~\cite{ma-etal-2020-simulmt, ren-etal-2020-simulspeech} is their end-to-end modeling of the whole process, which improves the efficiency compared to a cascade approach.
Such an end-to-end SimulST model is trained using speech translation corpora such as MuST-C~\cite{di-gangi-etal-2019-must}, but these corpora are usually based on offline translation due to the lack of large-scale SI data.

For the English-Japanese language pair, there have been some attempts for the development of SI corpora~\cite{Tohyama-et-al-2004, shimizu2013constructing, doi-etal-2021-large}.
However, the amount of such SI corpora is still very limited compared to offline translations.
We tackle this problem by using a larger-scale offline translation corpus.
This condition can be seen as domain adaptation from resource-rich offline translation to resource-poor simultaneous translation.
In a typical domain adaptation scenario, an out-of-domain model is fine-tuned using in-domain data~\cite{luong-manning-2015-stanford, sennrich-etal-2016-controlling}, but it tends to overfit to the small in-domain data~\cite{chu-etal-2017-empirical}.
As another adaptation approach, tag-based NMT works to control the politeness of translations ~\cite{sennrich-etal-2016-controlling} and to enable zero-shot multilingual NMT~\cite{johnson-etal-2017-googles}. 
This tag-based approach has been extended to multi-domain fine-tuning~\cite{kobus-etal-2017-domain} and mixed fine-tuning~\cite{chu-etal-2017-empirical}.
These studies fine-tune NMT models using mixed data of in-domain and out-of-domain corpora.
Tagged Back-Translation \cite{caswell-etal-2019-tagged} is an application of the tag-based approach to well-known back-translation-based data augmentation.
It distinguishes source language sentences from parallel corpora and those obtained from back-translation to handle possible back-translation noise in the training of an NMT model.
Our work is motivated by these tag-based methods and tackles the scarcity of SI data.

\section{Differences between Offline Translation and Simultaneous Interpretation}
There is a large style difference between SI and offline translation.
Figure~\ref{tab:example-sentence-in-si-off} shows an example of offline translation and SI transcript in Japanese for a given English source sentence.
The solid lines in the figure represent word correspondences.
In this figure, we can find:
\begin{itemize}
    \item Most English content words are translated into Japanese in the offline translation, while some are missing in the SI transcript.
    \item The SI tries to translate the former half of the input earlier than the latter half with some unnaturalness, while the offline translation keeps naturalness in Japanese with long-distance reordering from the input English.
\end{itemize}
These points suggest important differences between offline translation and SI;
SI focuses on the simultaneity of the interpretation to deliver the contents as early as possible and to maintain the interpreter's working memory.
The word order difference between English and Japanese poses a serious difficulty in SI, as mentioned in the literature \cite{mizuno2017simultaneous}.
Thus, it is important to use SI data to train a SimulST model to improve its simultaneity.

\section{Proposed Method}
\label{method}
Although training a SimulST model using SI data is necessary, we suffer from data scarcity in practice.
We propose a method to use a relatively large offline translation corpus to mitigate for the SI data scarcity for training a SimulMT model.
Following the tag-based NMT studies, we put a style tag at the beginning of the target string in training and predict a specified tag forcibly at the first step in inference.
In this work, we use two tags: \texttt{<si>} for SI and \texttt{<off>} for offline translation.

Suppose we have an SI transcript: 私は、買った。ペンを、 for an English input: \emph{I bought a pen.} as a training example.
We put the SI-style tag at the beginning of the SI transcript as follows:
\begin{quote}
\hspace{-4mm}\texttt{<si>}私は、買った。ペンを、
\end{quote}
This string is tokenized into subwords\footnote{``\_'' is the meta-character representing white spaces in an original string by SentencePiece~\cite{kudo-richardson-2018-sentencepiece}, and ``\verbvisiblespace'' represents a white space in a tokenized string.}:
\begin{quote}
\hspace{-4mm}\_\texttt{<}\verbvisiblespace\texttt{si}\verbvisiblespace\texttt{>}\verbvisiblespace私は\verbvisiblespace、\verbvisiblespace買っ\verbvisiblespaceた\verbvisiblespace。\verbvisiblespaceペン\verbvisiblespaceを\verbvisiblespace、
\end{quote}
Here, we assume we have a pre-trained sequence-to-sequence model such as mBART~\cite{liu-etal-2020-multilingual-denoising, tang2021multilingual} as a basis of the SimulST model, as described later in the next section.
The aforementioned style tags may not be included in the subword vocabulary of the pre-trained model and are tokenized further like ``\_\texttt{<}\verbvisiblespace\texttt{si}\verbvisiblespace\texttt{>}'', but it works in practice.

\section{Experimental Setup}
\subsection{Dataset}

\begin{table}[t]
    \small
    \centering
    \begin{tabular}{l|cc|cc}
    \hline
      & \multicolumn{2}{c|}{Offline} & \multicolumn{2}{c}{SI} \\
      & \#segm. & \#En words & \#segm. & \#En words \\
    \hline
        train & 328,639 & 5,714,360 & 65,008 & 1,120,245 \\
        dev & 1,369 & 23,059 & 165 & 2,804 \\
        test & 2,841 & 46,144 & 511 & 8,104\\ 
    \hline
    \end{tabular}
    \caption{Data sizes of offline data and SI data in the number of aligned segments. }
    \label{tab:data_size}
\end{table}

We used MuST-C~\cite{di-gangi-etal-2019-must} v2 English-Japanese data as our offline speech translation corpus.
We also prepared development and test sets from our in-house Japanese SI recordings on TED Talks that are not included in the training sets above. 
As for the SI data for training, we used NAIT-SIC-Aligned~\cite{zhao2023naist}. This SI data is constructed by applying heuristic sentence alignment to extract parallel sentence pairs using the latest version of NAIST-SIC\footnote{\url{https://dsc-nlp.naist.jp/data/NAIST-SIC/2022}}~\cite{doi-etal-2021-large}. From NAIST-SIC-Aligned, we selected INTRA, AUTO-DEV and AUTO-TEST as train, dev and test data, respectively.
For all the SI sets, we aligned the English text segments with the corresponding audio tracks in MuST-C using an English forced-aligner Gentle\footnote{\url{https://github.com/lowerquality/gentle}}.
Here, we excluded segments not aligned with the source speech from the aligned dataset.
Table~\ref{tab:data_size} shows the size of the offline and SI data. 

\subsection{Simultaneous Speech Translation}
We used our SimulST implementation based on \texttt{fairseq}~\cite{ott-etal-2019-fairseq}.
It followed the system architecture of the best-scored system in the IWSLT 2022 evaluation campaign~\cite{polak-etal-2022-cuni}, which used an offline ST model in the online simultaneous decoding based on Local Agreement (LA)~\cite{liu2020lowlatency}\footnote{We also tried wait-k~\cite{ma-etal-2019-stacl}, but LA worked better than wait-k in our pilot test.}.

\subsubsection{Offline ST Model}
We built the initial offline ST model by connecting two pre-trained models.
Firstly, we used HuBERT Large as the encoder, which consists of a feature extractor trained on 60k hours of unlabeled speech data Libri-Light \cite{librilight} and a transformer encoder layer.
The feature extractor is a 7-layer convolutional layer with a kernel size of (10,3,3,3,3,2,2), a stride of (5,2,2,2,2,2,2), and 512 channels, while the transformer encoder layer consists of 24 layers.
Next, we used the decoder portion of mBART50, an encoder-decoder model pre-trained with 50 language pairs, as the decoder.
The decoder consists of 12 layers of transformer decoders, and the embedding layer and linear projection weights are shared, with a size of 250,000.
The dimension of each layer of the transformer encoder and decoder is 1024, the dimension of the feed forward network is 4096, the number of multi-heads is 16, the activation function is the ReLU function, and the normalization method is pre-layer normalization~\cite{DBLP:conf/iclr/BaevskiA19}.
These two models are connected by an Inter-connection~\cite{2023-inter-connection} that weights each transformer layer of the encoder and integrates the output tensors of each layer in a weighted sum, and a length adapter \cite{tsiamas-etal-2022-pretrained}.
The length adapter is a 3-layer convolutional network with 1024 channels, the stride of 2, and the activation function of GELU.

The inputs are waveforms with a 16-kHz sampling rate that are normalized to zero mean and unit variance.
During training, each source audio is augmented \cite{wavaugment2020} with a probability of 0.8.
We train the model on MuST-C \cite{di-gangi-etal-2019-must}, CoVoST-2 \cite{wang-etal-2020-covost}, Europarl-ST \cite{Europarl2020}, and TED-LIUM \cite{Rousseau2012TEDLIUMAA}.
We use gradient accumulation and data parallelism to achieve a batch size of approximately 32 million tokens. 
We use Adam with $\beta_1=0.99$, $\beta_2=0.98$, and a base learning rate of $2.5\times10^{-4}$. 
The learning rate is controlled by a tri-stage scheduler with phases of 0.15, 0.15, and 0.70 for warm-up, hold, and decay, respectively, while the initial and final learning rate has a scale of 0.01 compared to base. 
We use sentence averaging and gradient clipping of 20. 
We apply a dropout of 0.1 before every non-frozen layer and use time masking for 10-length spans with a probability of 0.2, and channel masking for 20-length spans with a probability of 0.1 in the encoder feature extractor's output. 
The loss is the cross-entropy loss with label smoothing of 0.2.
We call this trained model \emph{base} model.

The \emph{base} model was fine-tuned using the offline training and development sets (Table \ref{tab:data_size}).
During fine-tuning, we set the learning rate of $2.5\times10^{-5}$, saved models in every 1,000 updates, and adopted checkpoint averaging over five-best checkpoints according to the loss on the development set.
We call this fine-tuned model \emph{base+O} model.
About those \emph{base} and \emph{base+O} models, we use the NAIST IWSLT 2023 Simultaneous speech-to-speech model for the Simultaneous Speech Translation task~\cite{fukuda2023iwslt-system}.
We further fine-tune the \emph{base+O} model using the SI data in the same manner to derive \emph{base+O+S} model.
Here, following ~\cite{tsiamas-etal-2022-pretrained}, to avoid overfitting the small SI data, the parameters of the following components were kept fixed: the feature extractor and feedforward layers of the encoder and the embedding, self-attention, and feedforward layers of the decoder.

\subsubsection{Fine-tuning using Prefix Alignment}
For further fine-tuning toward SimulST, we extracted prefix-to-prefix translation pairs from the available training sets using Prefix Alignment (PA)~\cite{kano-etal-2022-simultaneous}.
PA uses an offline translation model to find prefix-to-prefix translation pairs that can be obtained as intermediate translation results using a given offline translation model.
Finally, we fine-tuned the \emph{base+O} model using the prefix pairs.

\subsubsection{Compared Methods}
We compared the following conditions on the final fine-tuning data:
\begin{description}
    \item[Offline FT] Fine-tuned using the prefix pairs from the offline data (baseline in offline).
    \item[SI FT] Fine-tuned using the prefix pairs from the SI data (baseline in SI).
    \item[Mixed FT] Fine-tuned using prefix pairs from both of the offline and SI data (baseline in mixed).
    \item[Mixed FT~+~Style] Fine-tuned using prefix pairs from both of the offline and SI data with the style tags (proposed method).
    \item[Mixed FT~+~Style~+~Up] The SI portions were upsampled in \textbf{Mixed FT~+~Style} to balance the data size between the offline and SI data (proposed method).
\end{description}
Here, the prefix pairs from the offline data were obtained using \emph{base+O} model,
and those from the SI data were obtained using the \emph{base+O+S} model.
The hyperparameter settings for the fine-tuning were the same as that for the \emph{base+O} model.

\begin{figure*}[t]
\centering
    \begin{minipage}[b]{0.48\linewidth}
        \centering
        \includegraphics[width=1.0\linewidth]{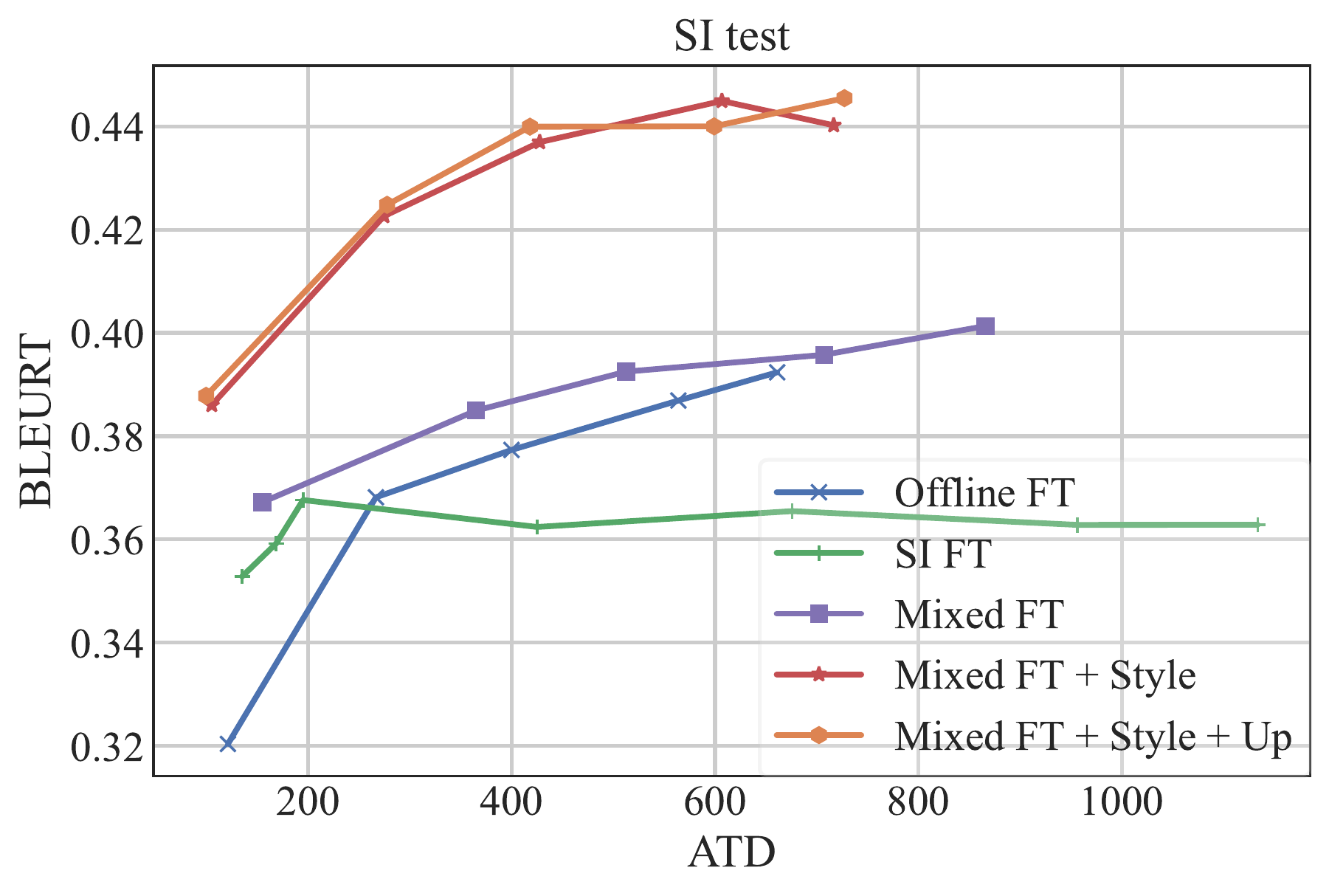}
        \subcaption{BLEURT}
        \label{fig:sharedtask-simulst-LA-test-xATD-yBLEURT}
    \end{minipage}
    \begin{minipage}[b]{0.48\linewidth}
        \centering
        \includegraphics[width=1.0\linewidth]{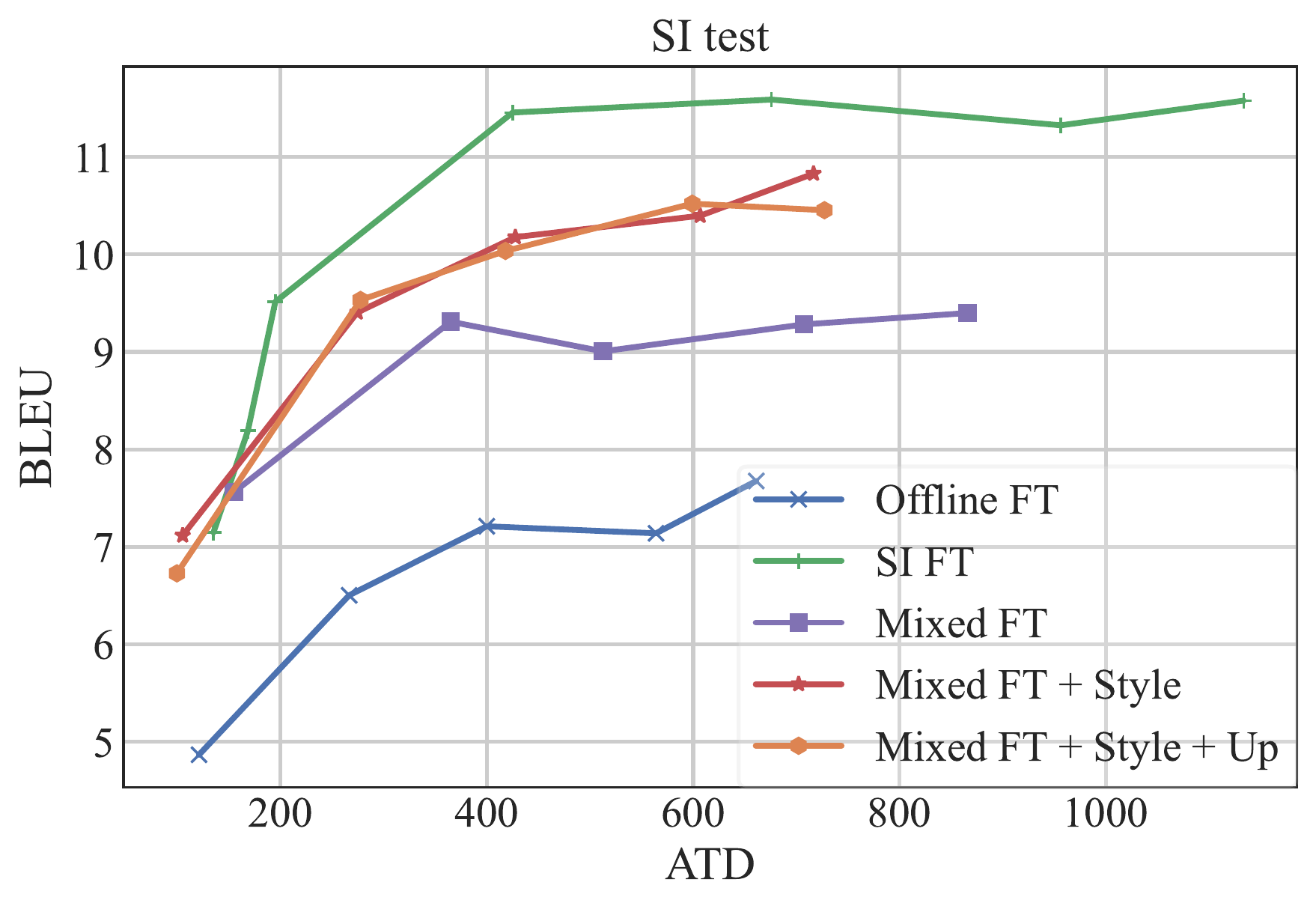}
        \subcaption{BLEU}
        \label{fig:sharedtask-simulst-LA-test-xATD-yBLEU}
    \end{minipage}
    \caption{SimulST latency (ATD) -- quality results on SI test set. }
    \label{fig:main-result-si-test}
    \begin{minipage}[b]{0.48\linewidth}
        \centering
        \includegraphics[width=1.0\linewidth]{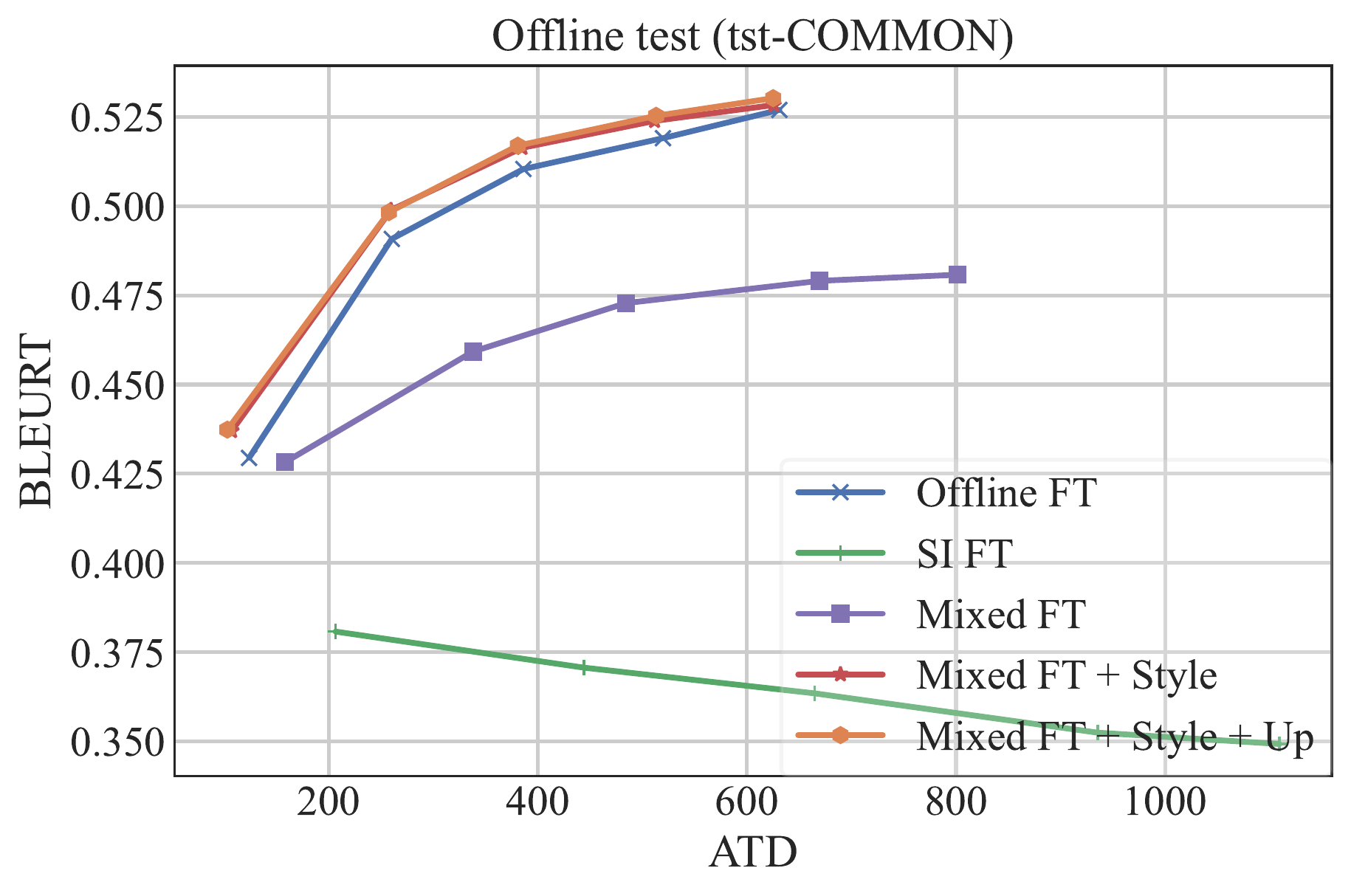}
        \subcaption{BLEURT}
        \label{fig:sharedtask-simulst-LA-tst-COMMON-xATD-yBLEURT}
    \end{minipage}
    \begin{minipage}[b]{0.48\linewidth}
        \centering
        \includegraphics[width=1.0\linewidth]{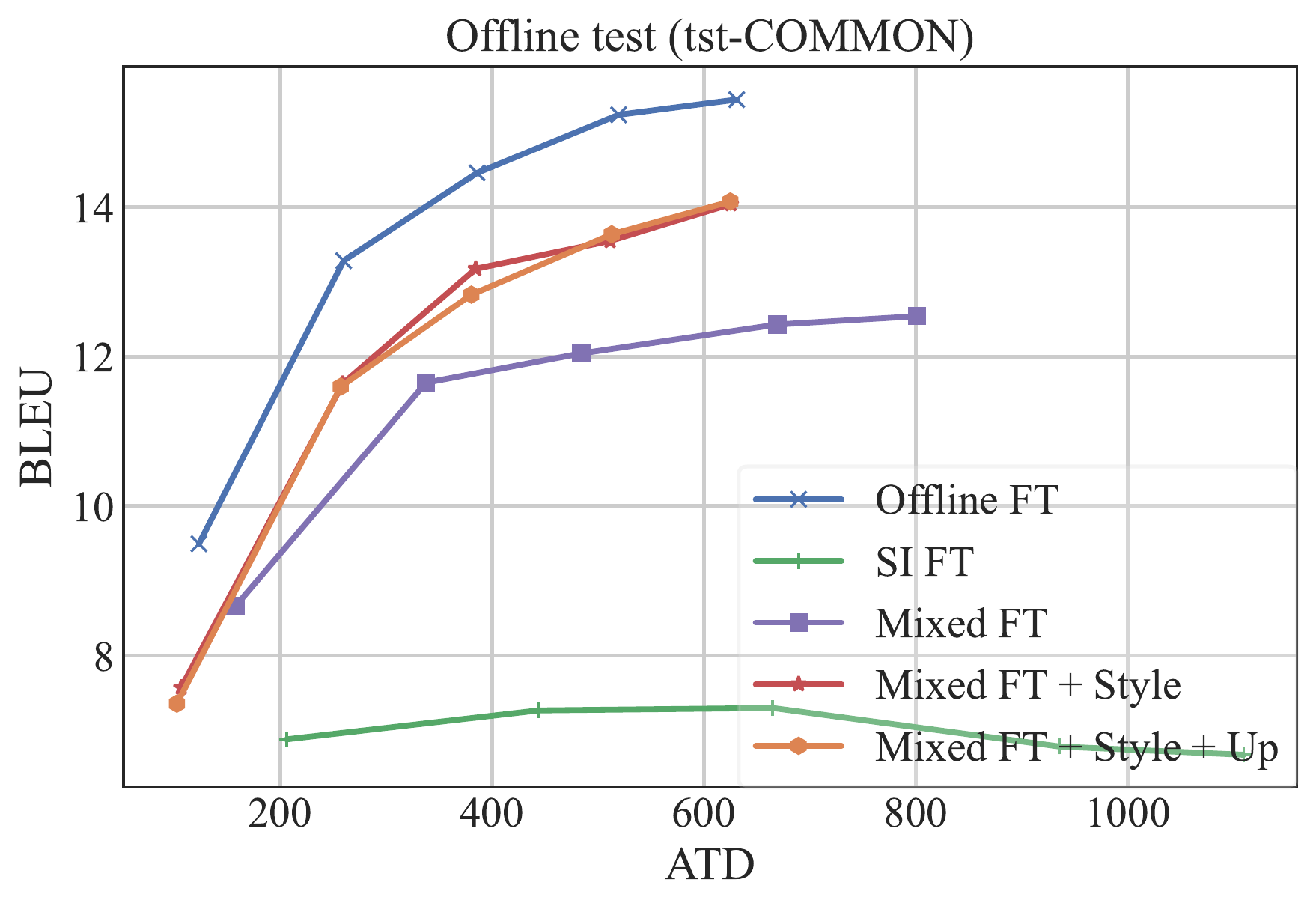}
        \subcaption{BLEU}
        \label{fig:sharedtask-simulst-LA-tst-COMMON-xATD-yBLEU}
    \end{minipage}
    \caption{SimulST latency (ATD) -- quality results on offline test set. }
    \label{fig:main-result-tst-COMMON}
\end{figure*}

\begin{table}[t]
    \centering
    \begin{tabular}{lcc}
    \hline
        (BLEURT) & SI & Offline \\ \hline 
        Offline FT & 0.386 & 0.518 \\
        SI FT & 0.359 & 0.347 \\
        Mixed FT & 0.393 & 0.483 \\ \hline
        Mixed FT + Style & \textbf{0.445} & \textbf{0.522}  \\ 
        Mixed FT + Style + Up & 0.443 & 0.516 \\ 
    \hline
    \end{tabular}
    \caption{BLEURT in full-sentence offline ST on SI and offline test sets. }
    \label{tab:offline-result-bleurt}
    \centering
    \begin{tabular}{lcc}
    \hline
        (BLEU) & SI & Offline \\ \hline 
        Offline FT & 7.8 & \textbf{16.0} \\
        SI FT & 10.9 & 6.3 \\
        Mixed FT & 9.4 & 13.3\\ \hline
        Mixed FT + Style & 10.3 & 15.4 \\ 
        Mixed FT + Style + Up & \textbf{12.2} & 14.2 \\ 
    \hline
    \end{tabular}
    \caption{BLEU in full-sentence offline ST on SI and offline test sets.}
    \label{tab:offline-result-bleu}
\end{table}

\subsection{Evaluation Metrics}
We evaluated the SimulST systems using SimulEval\footnote{\url{https://github.com/facebookresearch/SimulEval}}~\cite{ma-etal-2020-simuleval}.
The unit length of speech segments was set to \{200, 400, 600, 800, 1,000\} milliseconds\footnote{We also evaluated SI FT on the SI test set with 120 and 160 ms speech segments to investigate its performance in low latency ranges.}.
For the SimulST systems, translation quality was evaluated in BLEURT~\cite{sellam-etal-2020-bleurt} and BLEU~\cite{papineni-etal-2002-bleu}\footnote{BLEU was calculated using SacreBLEU~\cite{post-2018-call}.}.
The latency in SimulST was evaluated in Average Token Delay (ATD) \cite{kano2023average-interspeech} implemented in SimulEval.
Even though Average Lagging (AL) \cite{ma-etal-2019-stacl} is the most popular latency metric, it sometimes resulted in negative values, as suggested by \newcite{kano2023average-interspeech}.
Thus, we present the results using ATD and include the AL results in Appendix~\ref{sec:appendix-in-al}.

\section{Results}

\subsection{Offline Translation Results}

Tables~\ref{tab:offline-result-bleurt} and \ref{tab:offline-result-bleu} show the offline translation results in BLEURT and BLEU for the SI and offline test sets.
These results show that our proposed Mixed~FT~+~Style and Mixed~FT~+~Style~+~Up surpassed baselines in BLEURT for SI test.
On the offline test set (MuST-C tst-COMMON), the performance of the proposed models was almost the same as Offline~FT. 
This suggests that our proposed method leads to outputs semantically close to SI references than the baseline.
Contrary, the SI~FT baseline surpassed the Mixed~FT~+~Style in BLEU.
The result shows that the upsampling worked for BLEU improvement for the SI test set in the offline translation condition.

\subsection{Simultaneous Translation Results}

\begin{figure*}[t]
    \begin{minipage}[b]{0.33\linewidth}
        \centering
        \includegraphics[width=\linewidth]{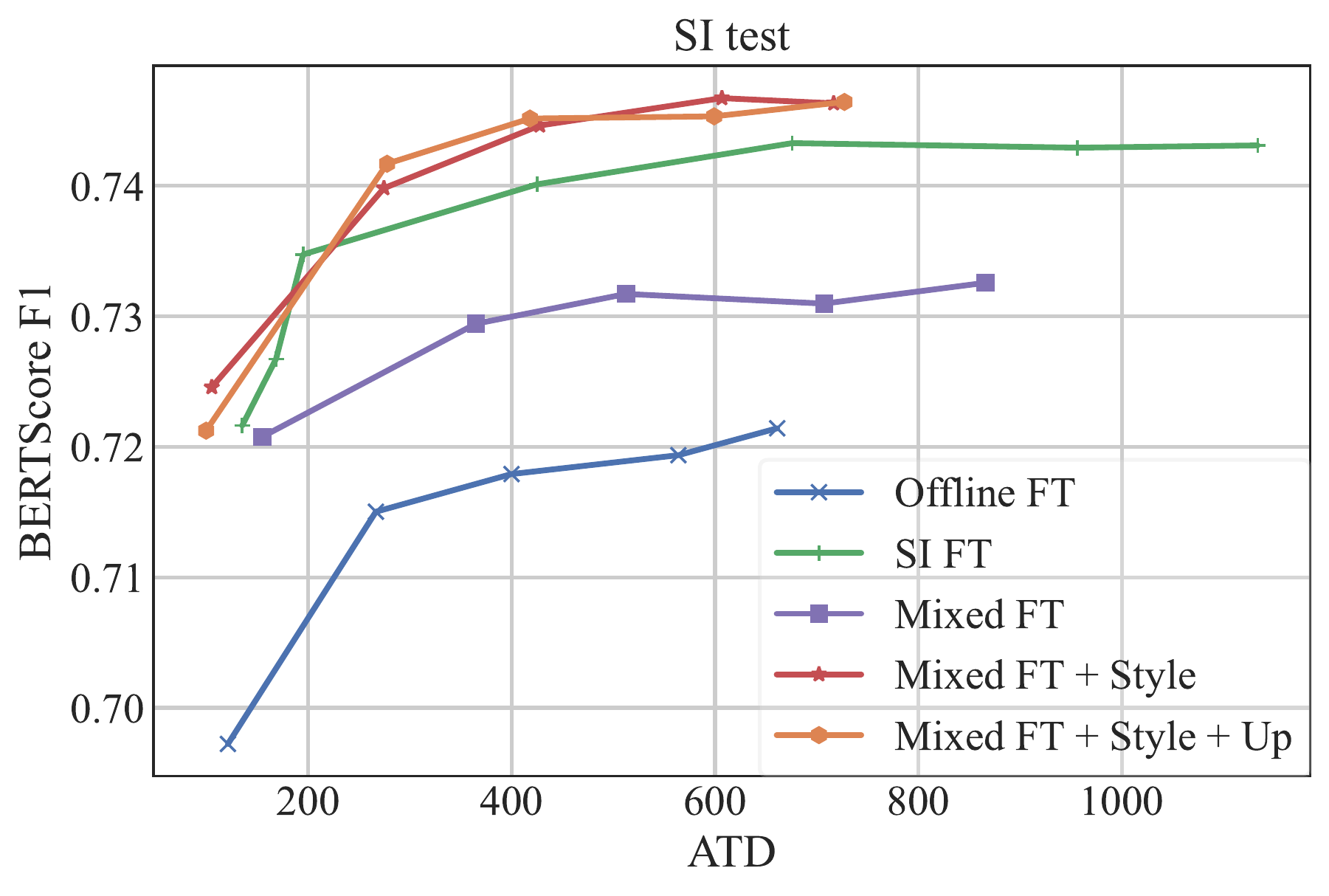}
        \subcaption{BERTScore F1}
        \label{fig:sharedtask-simulst-LA-test-xATD-yBERTScoreF1}
    \end{minipage}    
    \begin{minipage}[b]{0.33\linewidth}
        \centering
        \includegraphics[width=\linewidth]{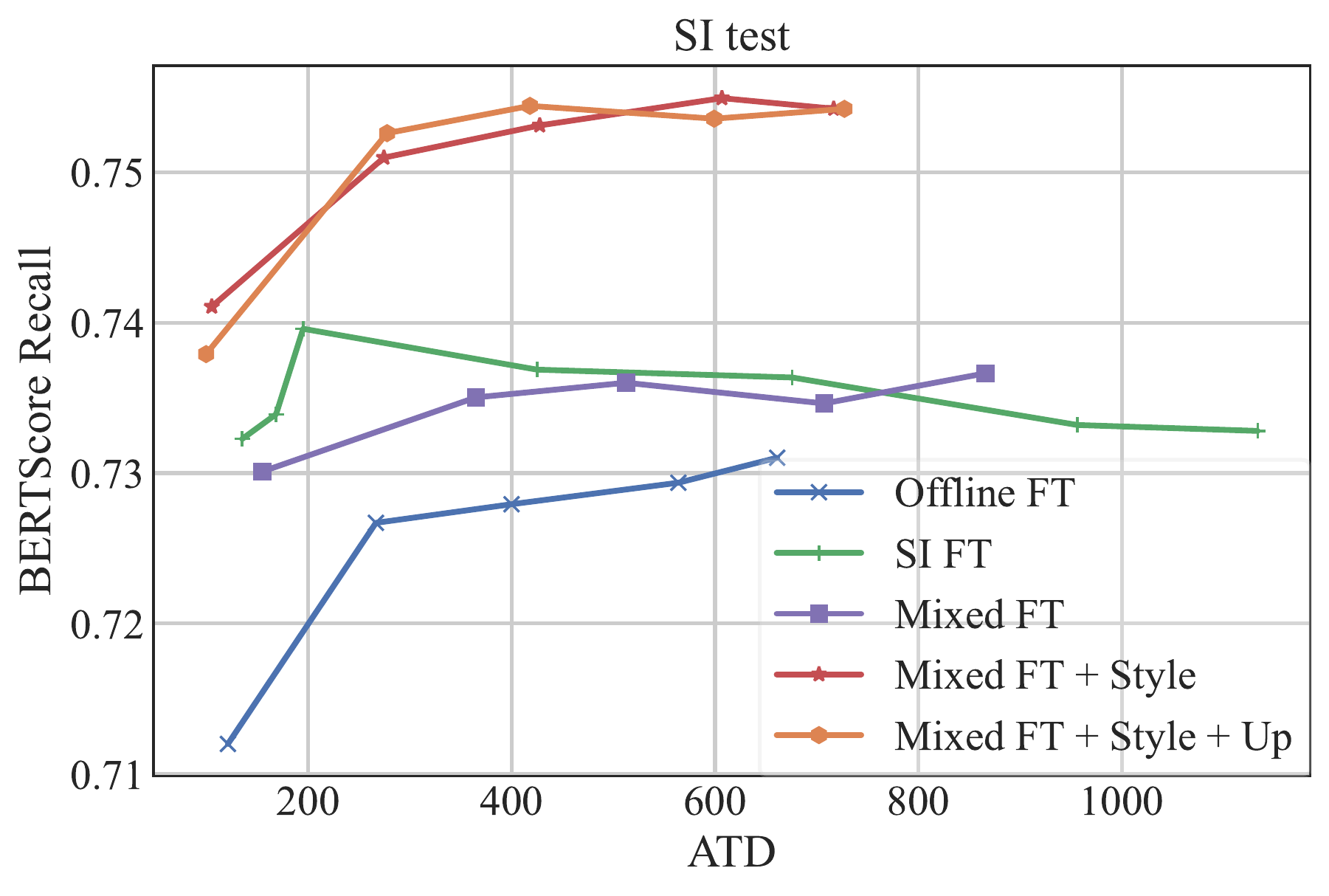}
        \subcaption{BERTScore Recall}
        \label{fig:sharedtask-simulst-LA-test-xATD-yBERTScoreF1Recall}
    \end{minipage}
    \begin{minipage}[b]{0.33\linewidth}
        \centering
        \includegraphics[width=\linewidth]{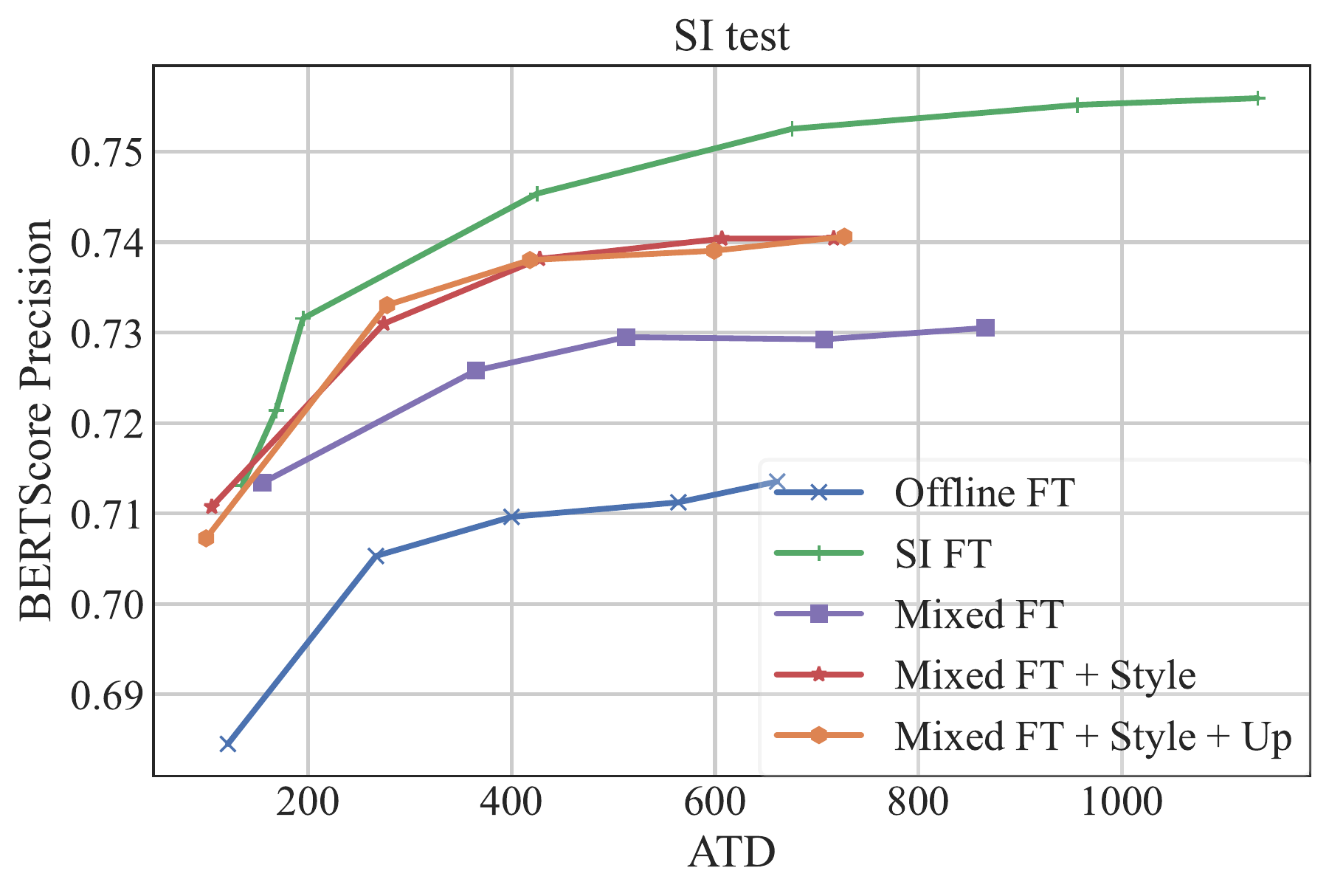}
        \subcaption{BERTScore Precision}
        \label{fig:sharedtask-simulst-LA-test-xATD-yBERTScorePrecision}
    \end{minipage}
    \caption{SimulST latency (ATD) -- quality (BERTScore) results on SI test set.}
    \label{fig:bertscore-all}
\end{figure*}

Figure~\ref{fig:main-result-si-test} shows SimulST results in BLEURT and BLEU for the SI test set.
In Figure~\ref{fig:sharedtask-simulst-LA-test-xATD-yBLEURT}, the proposed method with the style tags showed clearly better BLEURT results than the baselines.
The upsampling did not bring clear differences, the same as findings on the offline translation results shown in Table~\ref{tab:offline-result-bleurt}. 
In contrast, Figure~\ref{fig:sharedtask-simulst-LA-test-xATD-yBLEU} shows SI~FT worked the best in almost all latency ranges, while the proposed method outperformed the other two baselines (Offline and Mixed).

Figure~\ref{fig:main-result-tst-COMMON} shows SimulST results for the offline test set.
They reflect the difference in reference translations between the SI and offline test sets.
The Offline~FT baseline worked well in BLEURT and outperformed the proposed method in BLEU.
The other baselines resulted in worse BLEURT and BLEU scores than the proposed method.

These results suggest the proposed method conveys the information given in source language speech better than the baselines.

\section{Discussions}
The results shown in Figures~\ref{fig:main-result-si-test}, \ref{fig:main-result-tst-COMMON} demonstrated the advantage of the proposed method in BLEURT, but not in BLEU.
In this section, we discuss the results in detail to reveal which model works the best from the viewpoint of SimulST.

\subsection{BERTScore Details}
Figure~\ref{fig:bertscore-all} shows the detailed results in F1, recall, and precision by BERTScore~\cite{DBLP:journals/corr/abs-1904-09675-bertscore} for the SI test set. 
The proposed method worked the best in BERTScore recall, and the recall curves look similar to BLEURT curves shown in Figure~\ref{fig:sharedtask-simulst-LA-test-xATD-yBLEURT}.
On the other hand, the SI~FT baseline worked the best in BERTScore precision, and the precision curves look very similar to the BLEU curves shown in Figure~\ref{fig:sharedtask-simulst-LA-test-xATD-yBLEU}.
We conducted further analyses below to investigate the mixed results in different quality metrics.

\subsection{Length Differences}
\begin{figure}[t]
    \centering
    \includegraphics[width=0.9\linewidth]{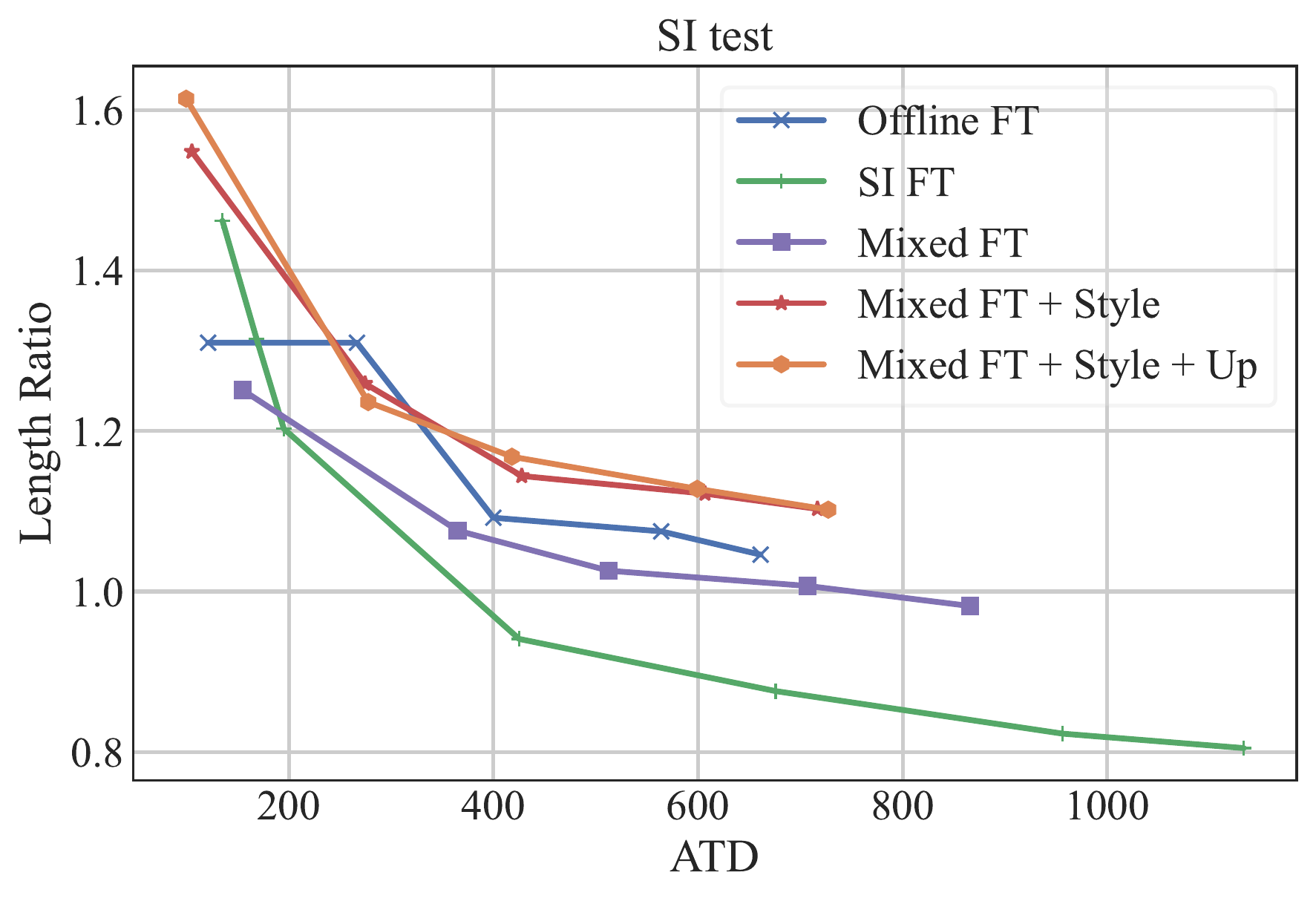}
    \caption{Length ratio results on SI test set.}
    \label{fig:sharedtask-simulst-xATD-yLengthRatio}

    \centering
    \includegraphics[width=0.9\linewidth]{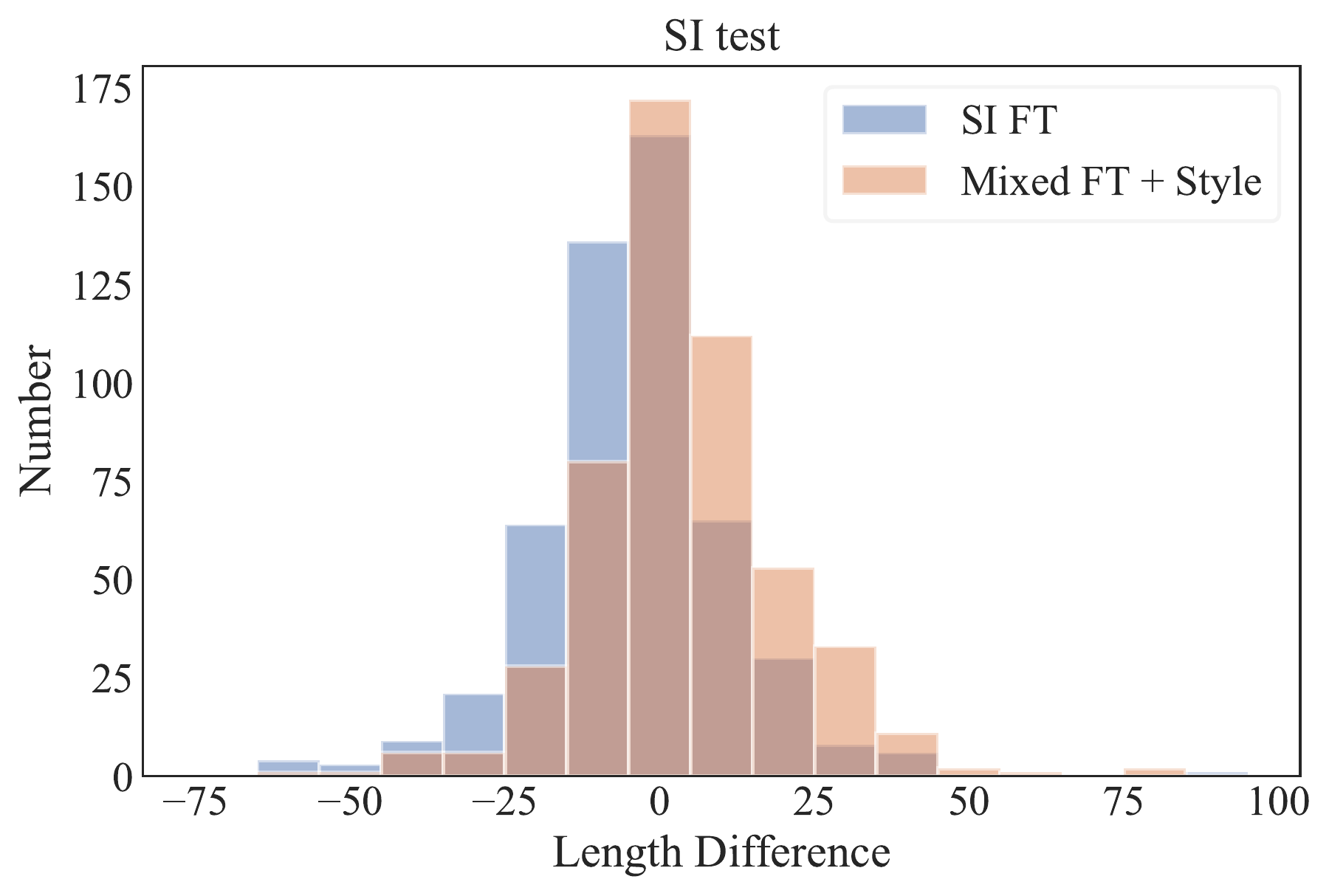}
    \caption{The length differences between hypotheses and references in SI FT and Mixed FT + Style (speech segment size is 600ms) on SI test set.}
    \label{fig:sharedtask-simulst-LA-test-histogram-delta-length}
\end{figure}

First, we focus on the length differences between translation outputs and references.
Figure~\ref{fig:sharedtask-simulst-xATD-yLengthRatio} shows the length ratios of translation results and their references.
The proposed method resulted in longer outputs than the baselines, and the SI~FT baseline preferred shorter output than the others and references.
From the viewpoint of the precision of the translation results, outputs longer than their references are unfavorable.
Figure~\ref{fig:sharedtask-simulst-LA-test-histogram-delta-length} shows the histogram of length differences between SI~FT and Mixed~FT~+~Style.
They showed different distributions; this suggests that SI~FT suffered from under-translation, and the proposed method suffered from over-translation.

\begin{table*}[t]
    \centering
    \small
    \begin{tabular}{ll}
    \hline
    Source   & TEMPT was one of the foremost graffiti artists in the 80s. \\
                & There's no hospital that can say ``No.'' \\
                & Anybody who's paralyzed now has access to actually draw or communicate using only their eyes. \\
    \hline
    SI FT & テンプトは、グラフィティアーティストの (\textit{TEMPT was, graffiti artists'}) \\
    (Baseline)  & 病院は、(\textit{a hospital}) \\
             & 麻痺した人達は、 (\textit{paralyzed people}) \\
    \hline
    Mixed~FT~+~Style & テンプトは、グラフィティアーティストの一人です。(\textit{TEMPT is one of graffiti artists.}) \\
    (Proposed)         & 病院では「いいえ」は言えません。(\textit{In a hospital, we cannot say ``No.''}) \\
             & 麻痺した人なら誰でも、絵を描いたり、会話をすることができます。\\
             & \qquad (\textit{Anybody who is paralyzed can draw a picture and have a talk.}) \\
    \hline
    SI reference & 八十年代の素晴らしいグラフィックアーティストでした。\\
                & \qquad (\textit{(He) was a great graphic artist in the 80s.})\\
              & 病院も、ノーとは言えない。(\textit{There's no hospital that can say ``No.''})\\
              & 麻痺してる人達は、これを全員使うことが出来るようになっています。 \\
              & \qquad (\textit{Everybody who is paralyzed can use this.})\\
    \hline
    Offline reference & 80年代を代表するグラフィティ・アーティストでした \\
              & 病院もダメと言えません \\
              & 全身麻痺の人誰もが 目だけで絵を描いたりコミュニケーションできます \\    
    \hline
    \end{tabular}
    \caption{Example sentences in SI~FT and Mixed~FT~+~Style (speech segment size: 600ms) on SI test set.}
    \label{tab:example-sentence-baseline-proposal-600ms}
\end{table*}

Table~\ref{tab:example-sentence-baseline-proposal-600ms} shows the translation examples by SI~FT and Mixed~FT~+~Style.
Here, SI~FT generates very short outputs compared with Mixed~FT~+~Style;
BLEU is not always good due to the brevity penalty, but SI~FT would have an advantage in BERTScore precision.

\subsection{Non-speech Sound Events and Repetitions}
Next, we investigated the over-translation suggested in the analyses above.

We observed serious repetitions by the proposed method, such as (拍手) (拍手) ..., which means (Applause).
This kind of non-speech sound events (applause and laughter) are found many times in TED Talks, but they are not translated by interpreters and excluded from the SI data.
According to this assumption, we tried to eliminate typical repetitions as follows and to conduct the evaluation after that.
\begin{itemize}
    \item Removing tokens if they are surrounded by "()" and "<>". (if the tokens include parts of "(拍手)" like "拍手)" or "(", they were also excluded.)
    \item Stopping the generating output when at least one kind of 3-gram appeared at least 3 times in the steps until reaching the end of the sentence. 
\end{itemize}

\begin{figure}[t]
\centering
    \begin{minipage}[b]{0.9\linewidth}
        \centering
        \includegraphics[width=\linewidth]{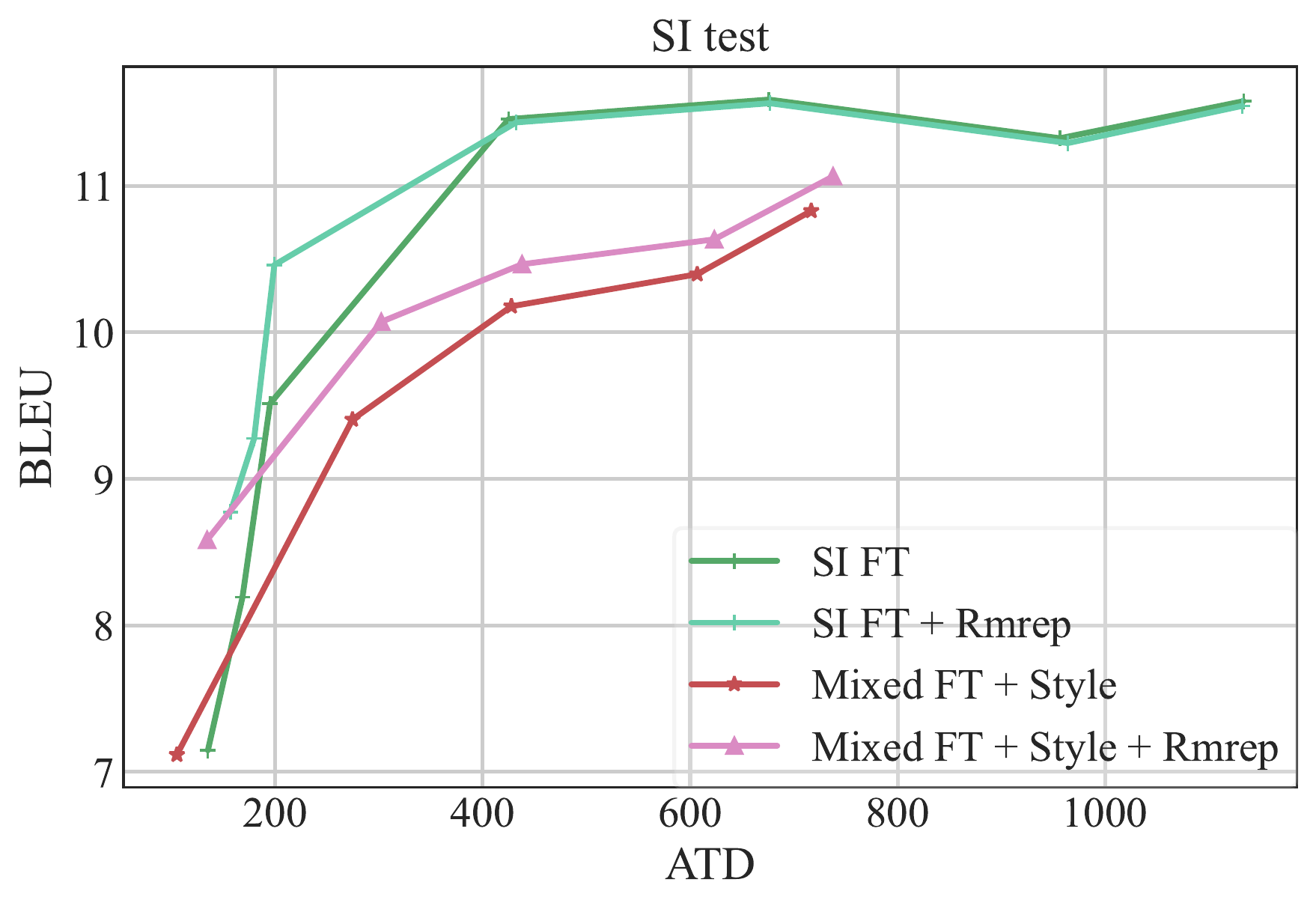}
        \subcaption{BLEU}
        \label{fig:sharedtask-simulst-LA-test-rmrep-bleu}
    \end{minipage}
    \begin{minipage}[b]{0.9\linewidth}
        \centering
        \includegraphics[width=\linewidth]{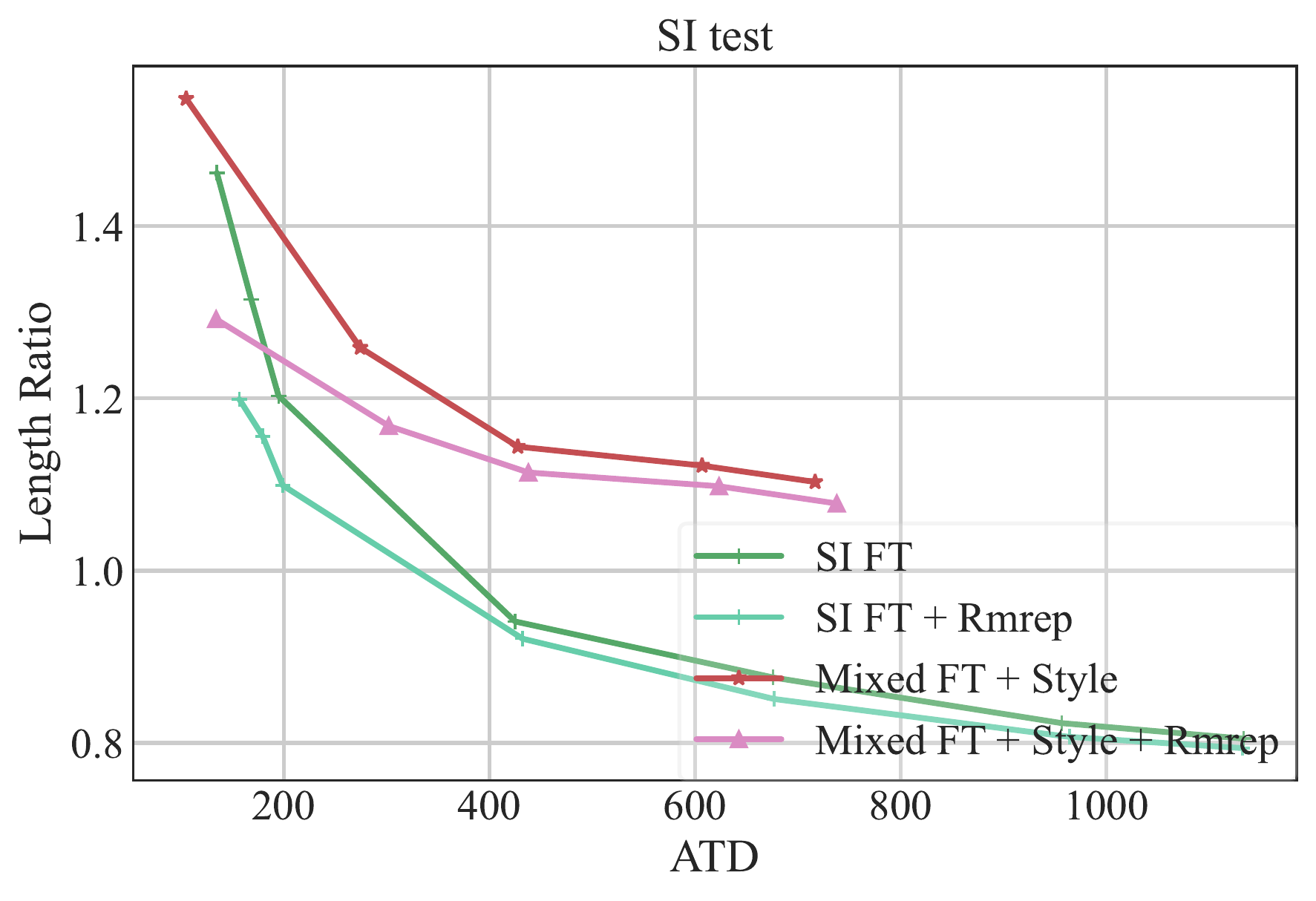}
        \subcaption{Length ratio}
        \label{fig:sharedtask-simulst-LA-test-rmrep-lengthratio}
    \end{minipage}
    \caption{Results with repetition removal (Rmrep) in BLEU and length ratio against ATD on SI test set. }
    \label{fig:sharedtask-simulst-LA-test-rmrep-bleu-lengthratio}
\end{figure}

\begin{figure}[p]
    \begin{minipage}[b]{0.9\linewidth}
        \centering
        \includegraphics[width=\linewidth]{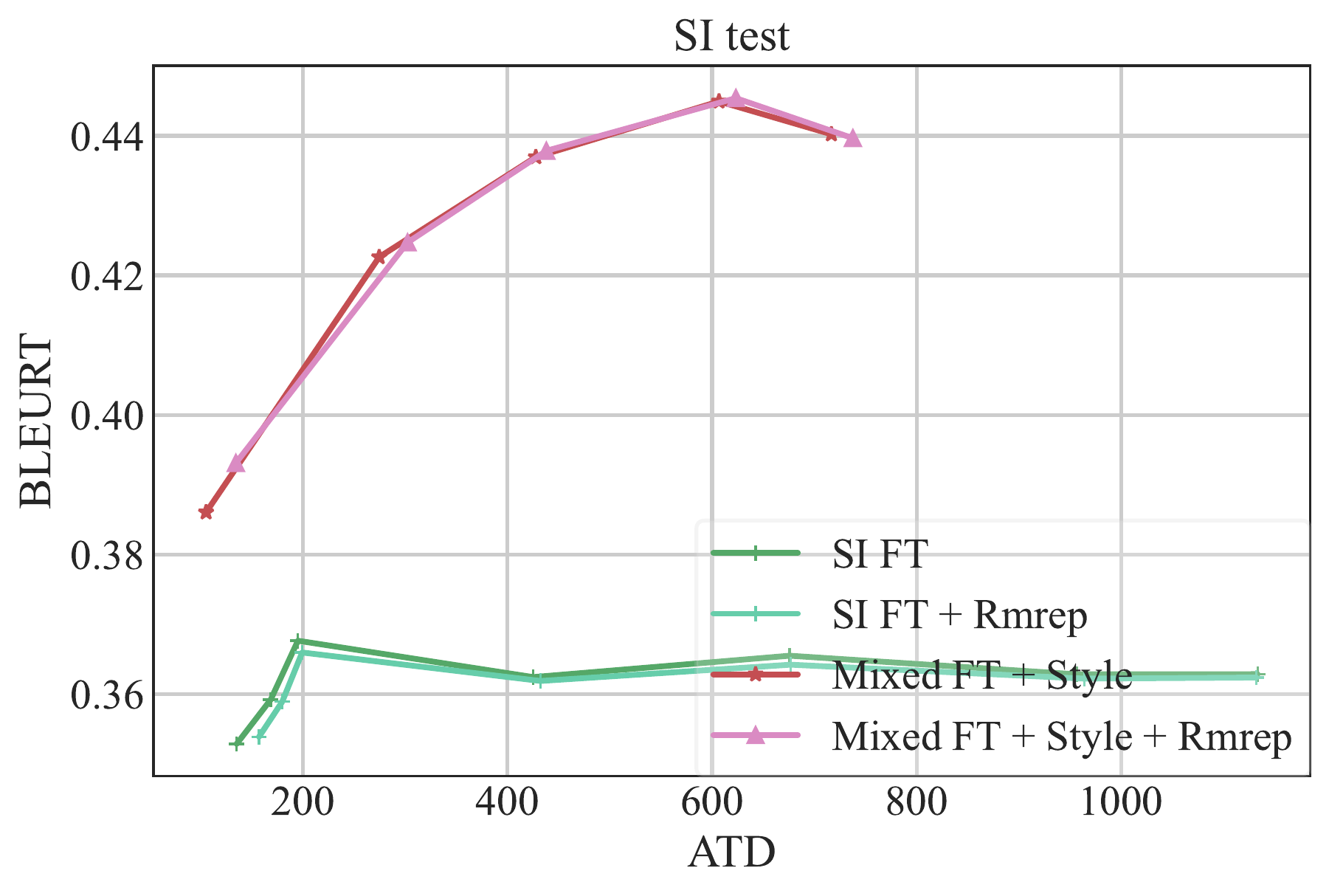}
        \subcaption{BLEURT}
        \label{fig:sharedtask-simulst-LA-test-rmrep-bleurt}
    \end{minipage}
    \begin{minipage}[b]{0.9\linewidth}
        \centering
        \includegraphics[width=\linewidth]{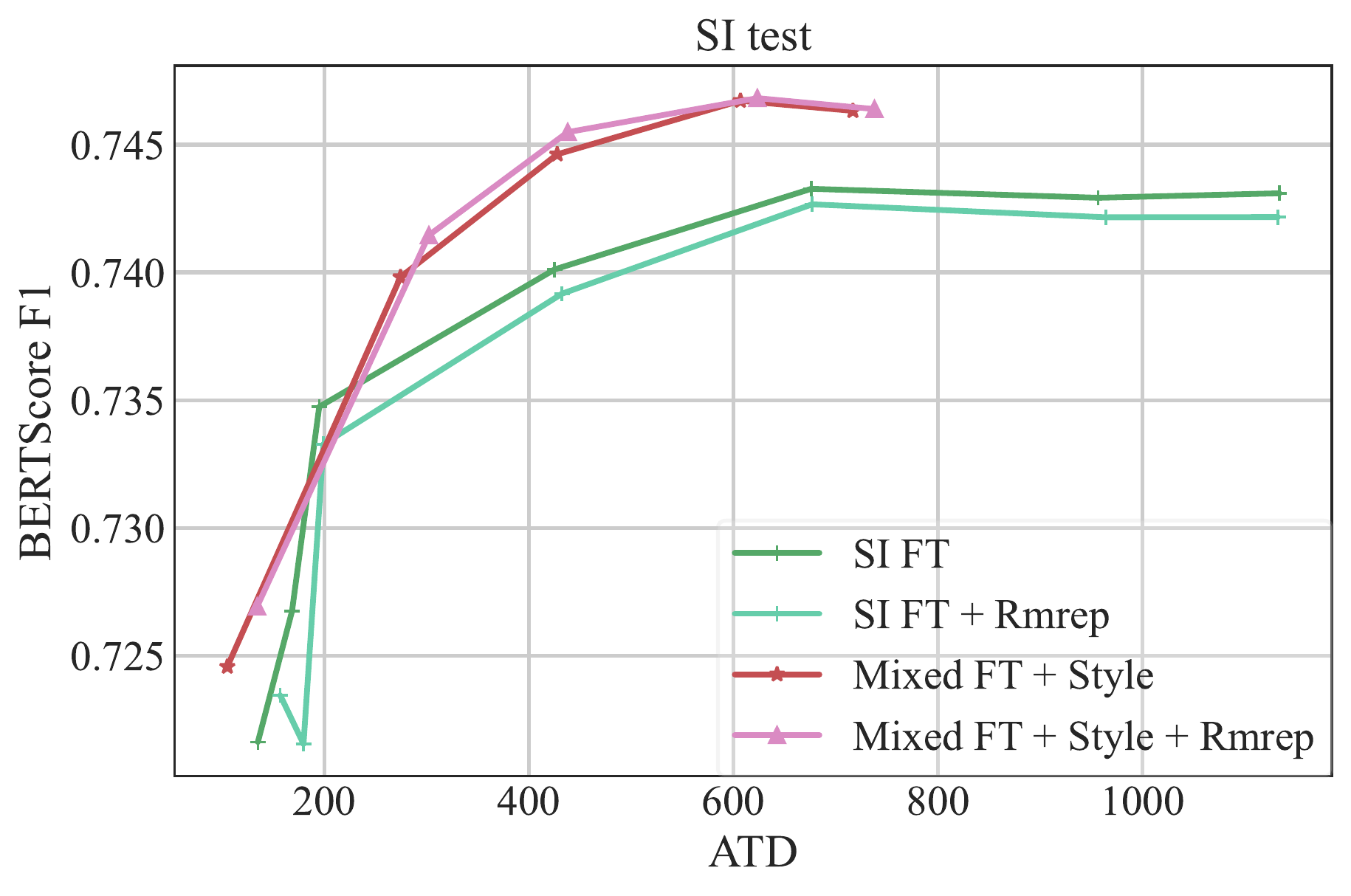}
        \subcaption{BERTScore-F1}
        \label{fig:sharedtask-simulst-LA-test-rmrep-bertscore-f1}
    \end{minipage}

    \begin{minipage}[b]{0.9\linewidth}
        \centering
        \includegraphics[width=\linewidth]{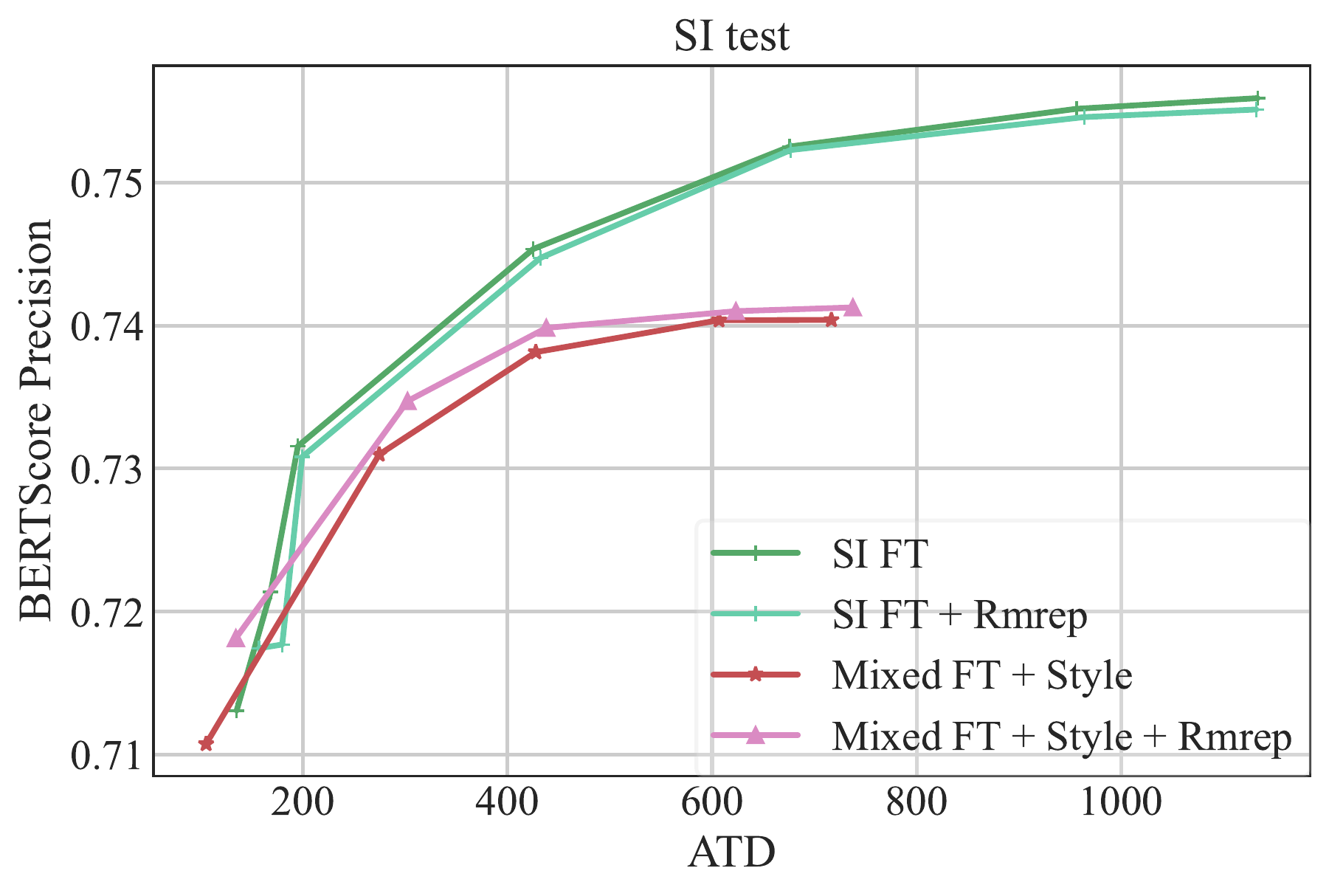}
        \subcaption{BERTScore-Precision}
        \label{fig:sharedtask-simulst-LA-test-rmrep-bertscore-precision}
    \end{minipage}
    \begin{minipage}[b]{0.9\linewidth}
        \centering
        \includegraphics[width=\linewidth]{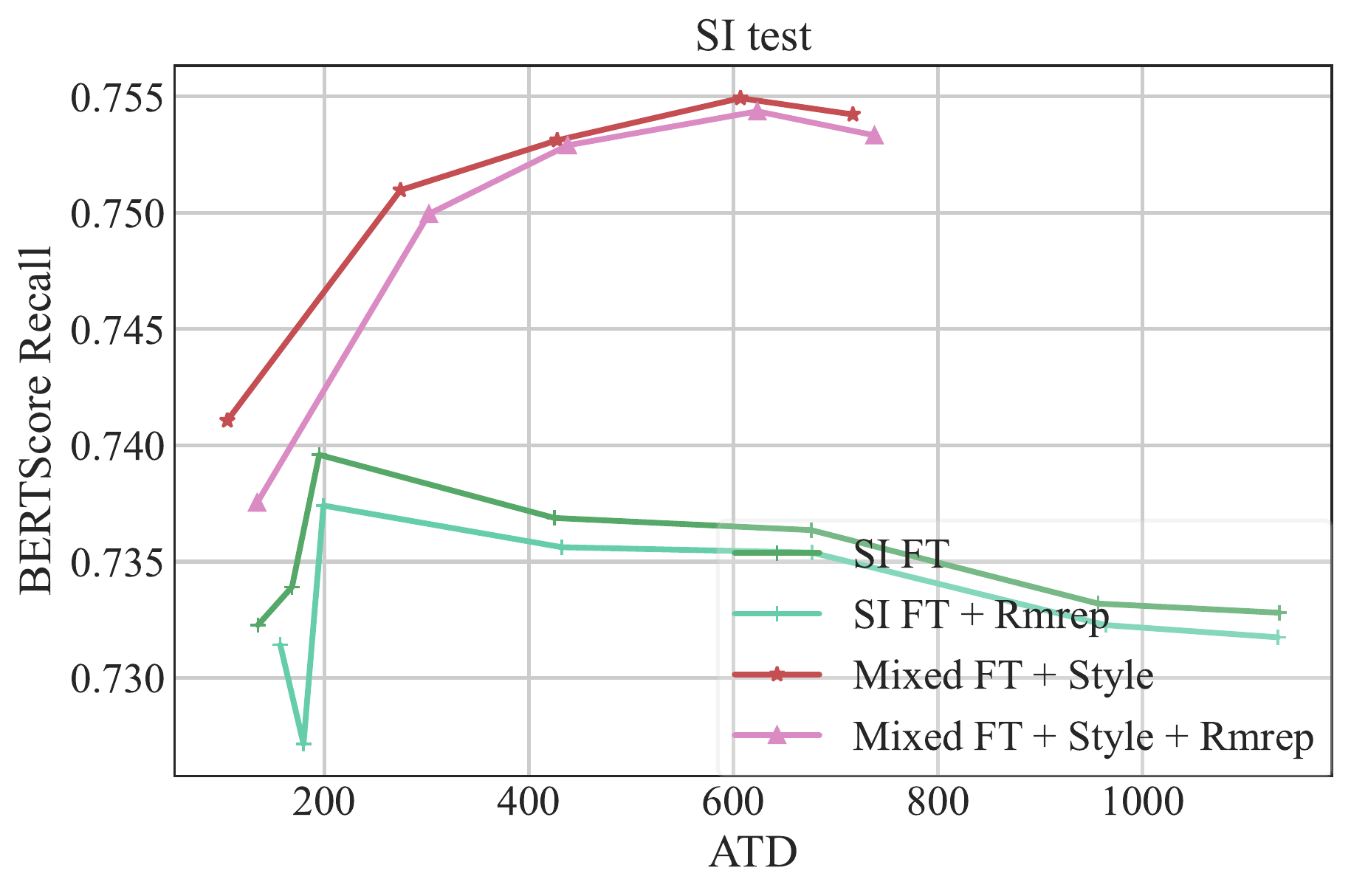}
        \subcaption{BERTScore-Recall}
        \label{fig:sharedtask-simulst-LA-test-rmrep-bertscore-recall}
    \end{minipage}
    \caption{Results with repetition removal (Rmrep) in BLEURT and BERTScore F1, precision and recall against ATD on SI test set. }
    \label{fig:sharedtask-simulst-LA-test-rmrep-bleurt-bertscore}
\end{figure}

We applied this repetition removal on the results by Mixed~FT~+~Style and SI~+~Style; they are labeled as Mixed~FT~+~Style~+~Rmrep and SI~FT~+~Rmrep, respectively.
Figure~\ref{fig:sharedtask-simulst-LA-test-rmrep-bleu-lengthratio} shows BLEU and length ratio results before and after the repetition removal.
BLEU increased consistently on the proposed method while almost no changes were observed on the SI~FT baseline except for one sample at ATD=200.
This suggests the existence of many repetitions in the translation results by the proposed method.
We also investigated BLEURT and BERTScore, as shown in Figure~\ref{fig:sharedtask-simulst-LA-test-rmrep-bleurt-bertscore}.
The repetition removal made almost no changes in BLEURT, probably due to the semantic-oriented evaluation strategy of BLEURT.
BERTScore Precision and F1 of the proposed method increased in the middle latency ranges, while they decreased almost consistently for the SI~FT baseline.
These findings suggest an over-translation problem with the proposed method, but it made little impact on semantic-oriented automatic evaluation results.

\section{Conclusion}
In this paper, we proposed an effective method to train a SimulST model using mixed data of SI- and offline-style translations with style tags to tell the model to generate outputs in either style, motivated by the tag-based approach to domain adaptation.
Experiment results on English-to-Japanese SimulST demonstrated the advantage of the proposed method in BLEURT and BERTScore recall despite the inferior performance in BLEU and BERTScore precision due to over-translations and repetitions.
Future work includes an extension to other language pairs and further verification via human evaluation.

\section{Limitation}
The scores reported in the SI test were lower than those in the offline test. 
Reporting results on other SI data would support seeing the effectiveness of our method. 
To our knowledge, this is the first work to use SI data as speech translation data. There are no other language pairs SI data than English-Japanese pairs those source speech and target text aligned. 

\section*{Acknowledgement}
Part of this work was supported by JSPS KAKENHI Grant Number JP21H05054 and JST SPRING Grant Number JPMJSP2140.

\bibliography{anthology,custom}
\bibliographystyle{acl_natbib}

\clearpage
\appendix
\section{Evaluation Results in AL.}
\label{sec:appendix-in-al}
\begin{figure*}[t]
\centering
    \begin{minipage}[b]{0.48\linewidth}
        \centering
        \includegraphics[width=\linewidth]{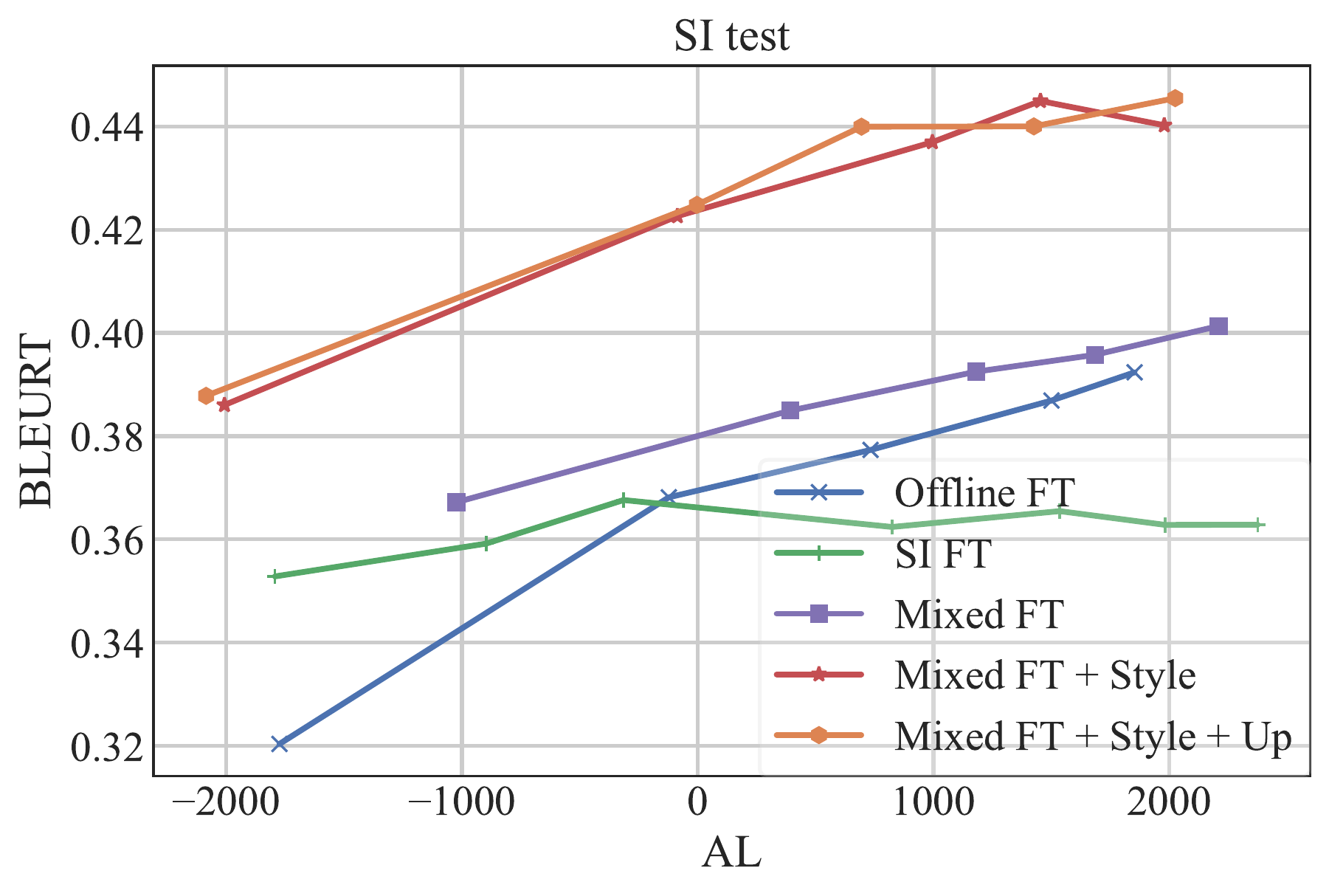}
        \subcaption{BLEURT}
        \label{fig:sharedtask-simulst-LA-test-xAL-yBLEURT}
    \end{minipage}
    \begin{minipage}[b]{0.48\linewidth}
        \centering
        \includegraphics[width=\linewidth]{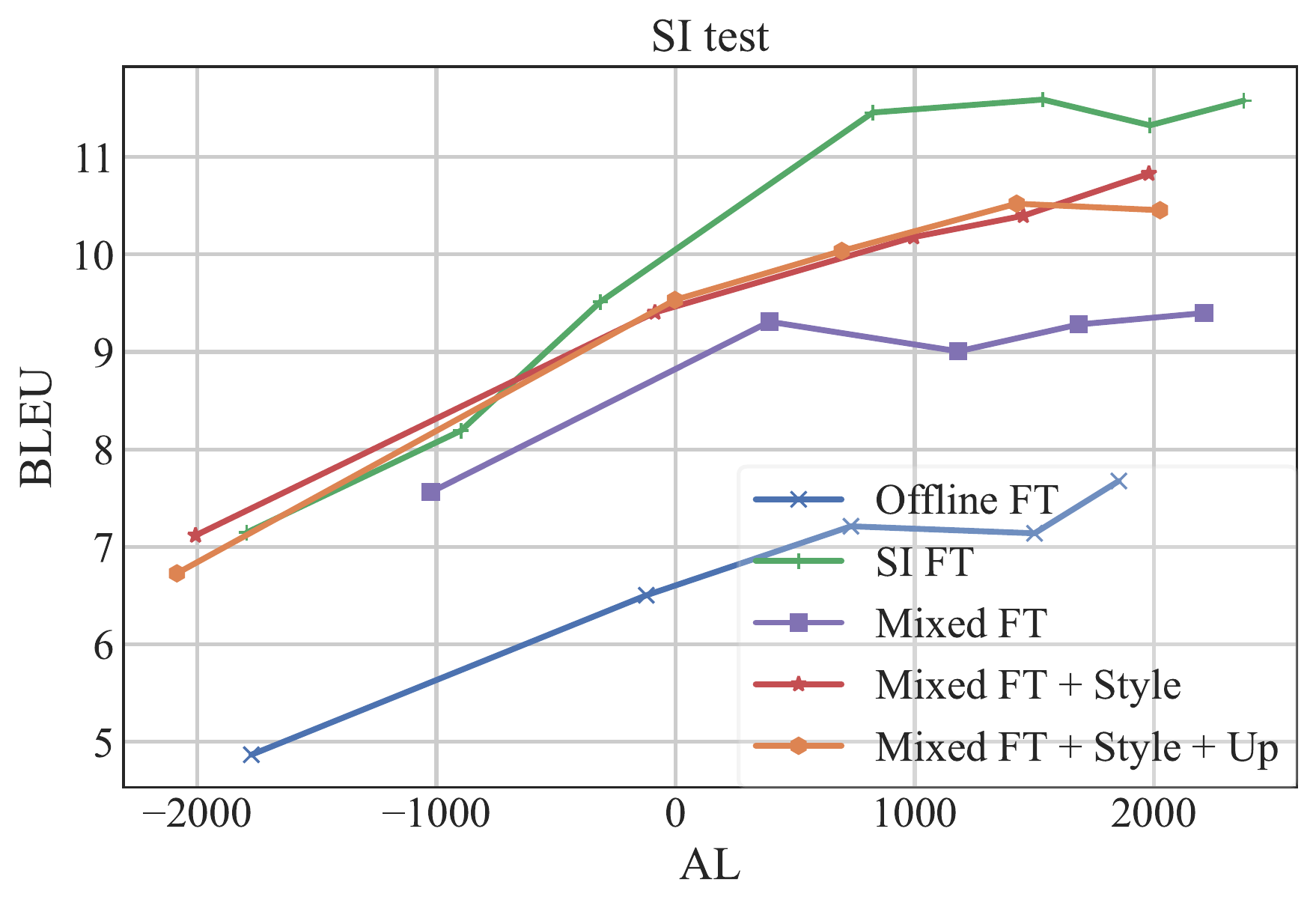}
        \subcaption{BLEU}
        \label{fig:sharedtask-simulst-LA-test-xAL-yBLEU}
    \end{minipage}
    \caption{SimulST latency (AL) – quality results on SI test set.}
    \label{fig:main-result-si-test-al}
    
    \begin{minipage}[b]{0.48\linewidth}
        \centering
        \includegraphics[width=\linewidth]{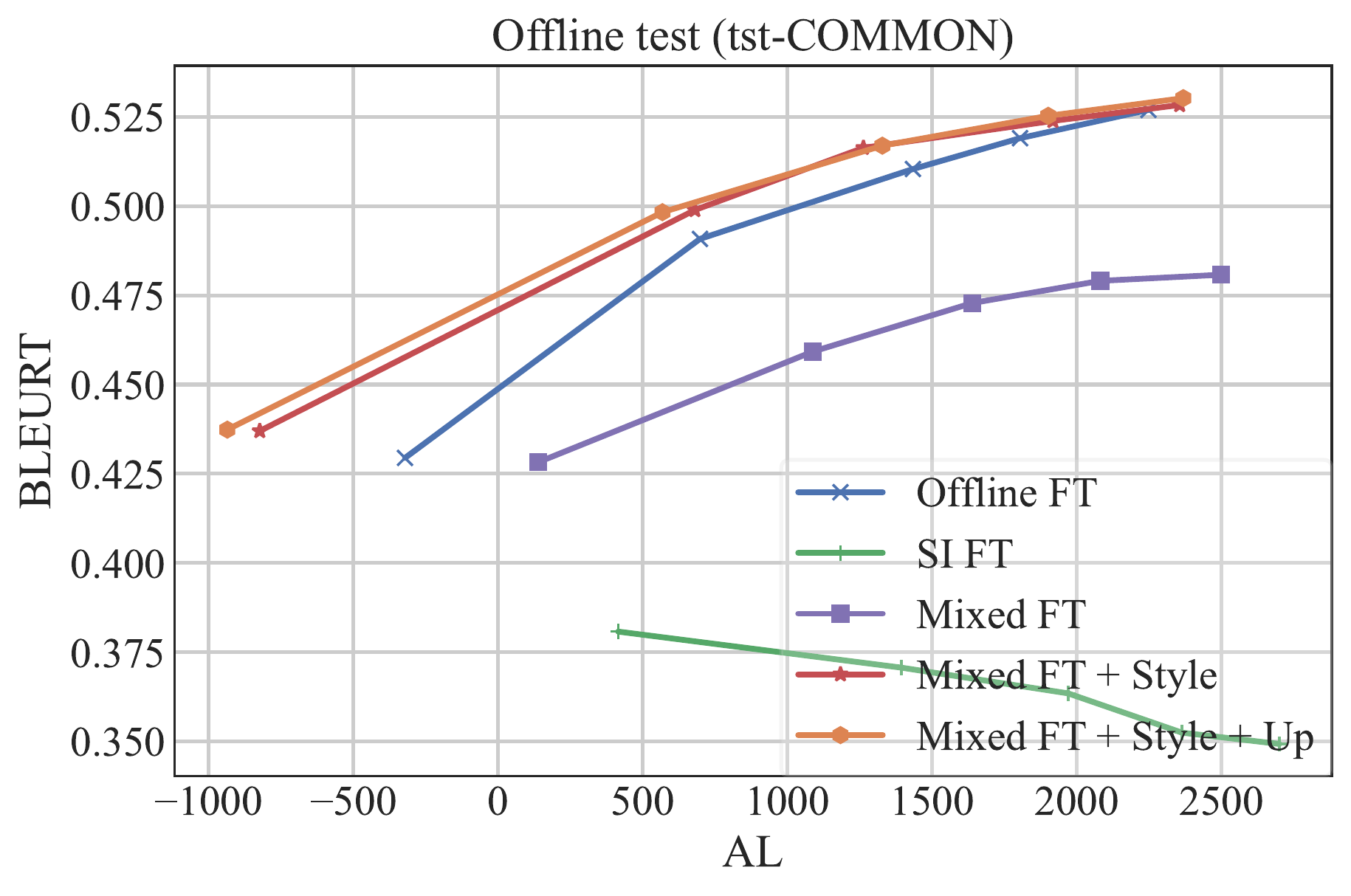}
        \subcaption{BLEURT}
        \label{fig:sharedtask-simulst-LA-tst-COMMON-xAL-yBLEURT}
    \end{minipage}
    \begin{minipage}[b]{0.48\linewidth}
        \centering
        \includegraphics[width=\linewidth]{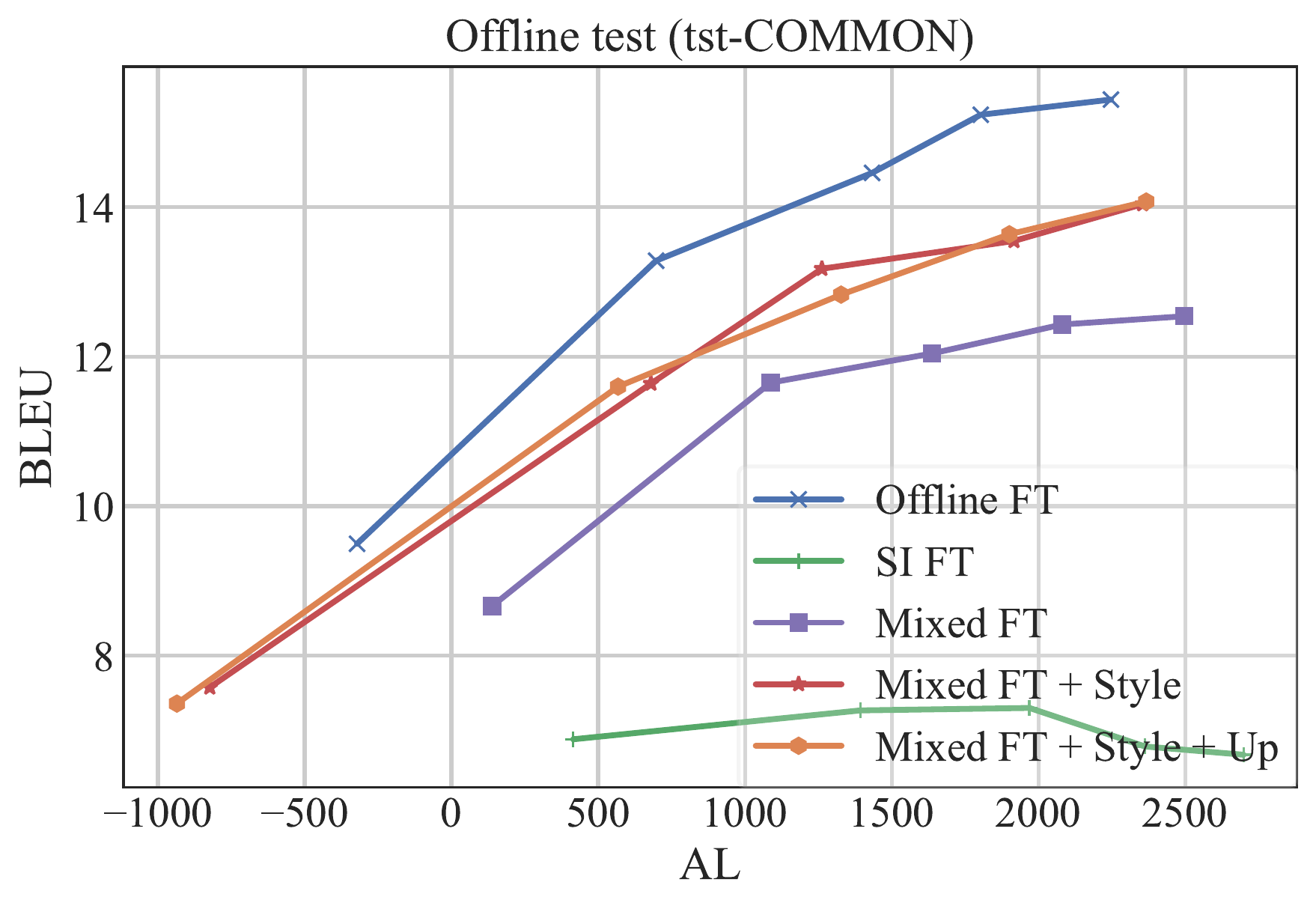}
        \subcaption{BLEU}
        \label{fig:sharedtask-simulst-LA-tst-COMMON-xAL-yBLEU}
    \end{minipage}
    \caption{SimulST latency (AL) – quality results on offline test set.}
    \label{fig:main-result-tst-COMMON-al}
\end{figure*}

Figure~\ref{fig:main-result-si-test-al} shows the main results in BLEURT and BLEU in SI test in AL.
Figure~\ref{fig:main-result-tst-COMMON-al} shows the main results in BLEURT and BLEU in offline test in AL.
Those results trends are almost the same as the trends in main results in Figure~\ref{fig:main-result-si-test}, \ref{fig:main-result-tst-COMMON}.

\end{document}